\documentclass[journal]{IEEEtran}

\usepackage{url}  
\usepackage{graphicx} 
\usepackage{amsmath,amssymb}
\usepackage{color}
\usepackage{multirow}
\usepackage{bbm}
\usepackage{amsfonts}
\usepackage{booktabs}
\usepackage{bm}
\usepackage{cite}
\usepackage{ifpdf}
\usepackage{booktabs}
\usepackage{threeparttable}
\usepackage[linesnumbered, boxed, ruled]{algorithm2e}

\usepackage[breaklinks=true,colorlinks,bookmarks=true]{hyperref}

\usepackage{bbding}

\usepackage[capitalize]{cleveref}
\crefname{section}{Sec.}{Secs.}
\Crefname{section}{Section}{Sections}
\Crefname{table}{Table}{Tables}
\crefname{table}{Tab.}{Tabs.}

\begin{document}

\title{SQLNet: Scale-Modulated Query and Localization Network for Few-Shot Class-Agnostic Counting}

\author{Hefeng Wu, Yandong Chen, Lingbo Liu, Tianshui Chen, Keze Wang, Liang Lin,~\IEEEmembership{Fellow,~IEEE}
\thanks{This work was supported by National Natural Science Foundation of China (NSFC) under Grant No. 62272494, 62325605 and 62306258, Guangdong Basic and Applied Basic Research Foundation under Grant No. 2023A1515012845 and 2023A1515011374, and Guangdong Province Key Laboratory of Big Data Analysis and Processing. (Corresponding author: Liang Lin)}
\thanks{Hefeng Wu, Yandong Chen, Keze Wang, and Liang Lin are with Sun Yat-sen Univeristy, Guangzhou 510275, China (e-mail: wuhefeng@gmail.com, chenyd35@mail2.sysu.edu.cn, kezewang@gmail.com, linliang@ieee.org).} 
\thanks{Lingbo Liu is with Peng Cheng Laboratory, Shenzhen 518055, China (e-mail: liulb@pcl.ac.cn).}
\thanks{Tianshui Chen is with Guangdong University of Technology, Guangzhou 510006, China (e-mail: tianshuichen@gmail.com).}%
}

\markboth{IEEE Transactions on Image Processing}%
{Wu \MakeLowercase{\textit{et al.}}: }

\maketitle

\begin{abstract}

The class-agnostic counting (CAC) task has recently been proposed to solve the problem of counting all objects of an arbitrary class with several exemplars given in the input image. To address this challenging task, existing leading methods all resort to density map regression, which renders them impractical for downstream tasks that require object locations and restricts their ability to well explore the scale information of exemplars for supervision. Meanwhile, they generally model the interaction between the input image and the exemplars in an exemplar-by-exemplar way, which is inefficient and may not fully synthesize information from all exemplars. To address these limitations, we propose a novel localization-based CAC approach, termed Scale-modulated Query and Localization Network (SQLNet). It fully explores the scales of exemplars in both the query and localization stages and achieves effective counting by accurately locating each object and predicting its approximate size. Specifically, during the query stage, rich discriminative representations of the target class are acquired by the Hierarchical Exemplars Collaborative Enhancement (HECE) module from the few exemplars through multi-scale exemplar cooperation with equifrequent size prompt embedding. These representations are then fed into the Exemplars-Unified Query Correlation (EUQC) module to interact with the query features in a unified manner and produce the correlated query tensor. In the localization stage, the Scale-aware Multi-head Localization (SAML) module utilizes the query tensor to predict the confidence, location, and size of each potential object. Moreover, a scale-aware localization loss is introduced, which exploits flexible location associations and exemplar scales for supervision to optimize the model performance. Extensive experiments demonstrate that SQLNet outperforms state-of-the-art methods on popular CAC benchmarks, achieving excellent performance not only in counting accuracy but also in localization and bounding box generation.

\end{abstract}


\IEEEpeerreviewmaketitle


\vspace{2ex}

\section{Introduction}

\IEEEPARstart{V}{isual} object counting underscores its practical importance with a fundamental and crucial role in a wide range of fields, encompassing daily business, security, and industrial production. However, the existing object counting paradigms are predominantly class-specific, confining the learned model solely to counting objects of a specific class, such as persons \cite{DDMD} or cars \cite{CARPK}. 
These class-specific counting paradigms exhibit inherent inflexibility, coupled with the imposition of substantial costs. Specifically, for application scenarios that are in the absence of a counting model tailored to a particular class, it becomes imperative to collect a large amount of annotated data to train a new model, thereby incurring significant expenses. 
Consequently, the practical applicability of such models is substantially curtailed in real-world scenarios.

\begin{figure}[!t]
    \centering
    \includegraphics[width=0.98\linewidth]{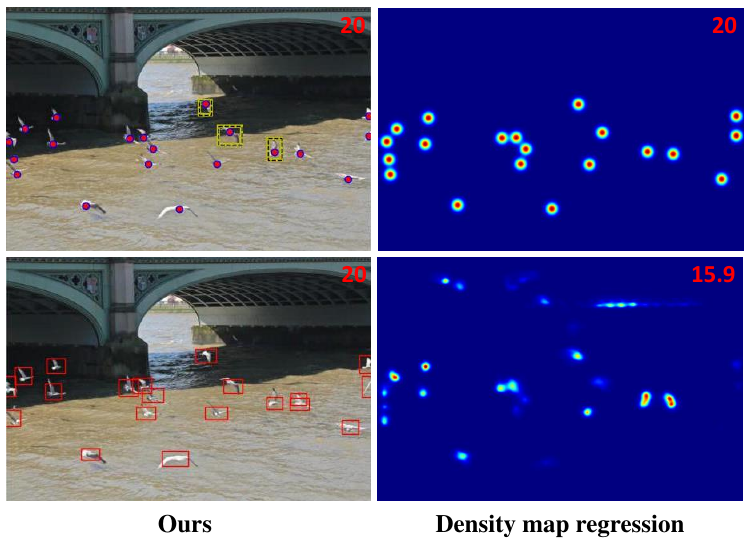}
    \caption{Illustration for the comparison between our method and density map regression based methods.  All previous state-of-the-art CAC methods follow a density map regression scheme. \textbf{Top row:} The left image shows the ground truth, where each targeted object is given by a point. The right image is a ground-truth density map generated from the points. The input is a raw image with several exemplars' bounding boxes given. \textbf{Bottom row:} The left image shows the result of our method, predicting each object's location along with its size. The right image shows the result of the leading method SAFECount \cite{SAFECount}. The top-right corner of an image exhibits the counting number. }
    \label{fig:intro}
\end{figure}

To address these issues, the task of class-agnostic counting (CAC) has recently emerged, aiming to count all objects of an arbitrary class in an image based on just a few provided exemplars.
Presently, all state-of-the-art CAC methods revolve around the utilization of density map regression \cite{bmnet,SAFECount}. These methods have demonstrated commendable performance, but there is still much room to improve the accuracy of counting.
Furthermore, this paradigm has limitations extending beyond counting accuracy concerns. In many real-world scenarios, it cannot meet the requirements of downstream tasks that require object locations, let alone the sizes of objects. 
It is intrinsically difficult to take the size into supervision. 
On the other hand, the object detection paradigm aims to predict the bounding boxes of objects. However, these methods require bounding box annotations as ground truth for supervision, while the counting task only provides location annotations. Previous works \cite{famnet,bmnet} have shown that adapting the detection paradigm \cite{fr,fsod} to the CAC task results in much poorer counting performance compared to the density regression paradigm. 
To overcome these limitations, we introduce a novel localization-based method to solve the CAC task, termed Scale-modulated Query and Localization Network (SQLNet).

Figure \ref{fig:intro} illustrates a visual comparison between our proposed method and the density map regression based state-of-the-art methods \cite{SAFECount,bmnet,famnet}. 
It can be observed that, by approximating the ground truth density map with hotspots representing the object locations (top-right image), the predicted density map (bottom-right image) can provide rough locations of some obvious objects. However, some sharp distributions also appear on the background, and many other objects are hard to discover due to smooth and flat density distributions, so the counting number needs to be obtained by summing up the density map.
In contrast, our method directly locates each object for counting, along with size prediction.
In the form of prediction output, it is akin to the detection paradigm.
However, the framework design and learning paradigm are different.
Specifically, we exploit a scale-aware localization loss, which fully harnesses flexible location associations and exemplar scales for supervision.
Our method can achieve excellent performance not only in counting accuracy but also in object localization and size prediction. 
To the best of our knowledge, we are the first to explore a localization-based method for the CAC task that achieves superior performance over the state-of-the-art methods based on density map regression.

Delving into the CAC task, due to the fact that exemplars are scarce and no prior information is available for an arbitrary class, one crucial foundation of a solid solution is to model the interaction between the query image and the exemplars effectively.
Existing state-of-the-art CAC methods obtain explicit similarity maps \cite{bmnet} or implicit correlation information \cite{famnet,SAFECount} from the interaction to perform density map regression. 
However, they generally model the interaction between the two in an exemplar-by-exemplar way, which is inefficient and may not comprehensively synthesize information from all exemplars.
It motivates us to investigate more effective ways to acquire sufficient correlation information. 
Therefore, in this work, we propose to accomplish the query stage of our framework from two aspects: (i) exploring multi-scale exemplars collaboration to obtain rich discriminative representations of the target class specified by the limited exemplars; (ii) conducting spatial and channel-wise interactions in an exemplars-unified manner. Specifically, we design two novel modules, i.e., Hierarchical Exemplars Collaborative Enhancement (HECE) and Exemplars-Unified Query Correlation (EUQC), to fulfill the above purpose.
Given the query image with few exemplars denoting the class of interest, the two modules will produce the correlated query tensor, which is used later for scale-aware localization.

In this work, we provide a new and effective solution to the CAC task. The main contributions of our work are summarized as follows:
\begin{itemize}
\item  We propose a novel framework SQLNet for the CAC task that achieves excellent accuracy not only in counting but also in localization and bounding box generation. To the best of our knowledge, we are the first to explore an explicit localization-based scheme that outperforms state-of-the-art CAC methods.
\item To capture sufficient correlation information between the input image and the exemplars in the query stage, we propose multi-scale exemplars collaboration with equifrequent size prompt and exemplars-unified spatial-channel interaction, and we introduce novel architecture designs to achieve them. 
\item To fully harness object locations and exemplar scales, we introduce a scale-aware localization learning paradigm. It adopts a flexible location matching strategy and exploits exemplar sizes for regularized supervision. It can facilitate the model to better focus on the target objects and lead to more precise and robust counting. 
\item Extensive evaluation and comparison with state-of-the-art methods are conducted on popular CAC benchmarks to verify the effectiveness of the proposed SQLNet and provide a comprehensive understanding of it. Our codes are available at \url{https://github.com/HCPLab-SYSU/SQLNet}.
\end{itemize}

\vspace{3ex}

\section{Related Work}
\label{sec:related}

In this section, we review the related works following two main research streams: class-specific counting and class-agnostic counting. 

\subsection{Class-Specific Counting}

Given any input image, class-specific counting aims at counting objects of a particular class in the image, such as people \cite{wang2021self,DDMD,crowdclip}, cars \cite{CARPK} and animals \cite{arteta2016counting}, etc. In this task, models are aware of the specific object class to be counted beforehand. They require individualized training for a specific object class, which consequently entails the collection of a large amount of annotated data and incurs heavy costs when dealing with each specific class.
Currently, existing approaches in the field can be broadly classified into three categories: detection-based, regression-based, and localization-based schemes. 

Detection-based methods \cite{CARPK,2011discriminative,sam2020locate, 2012detection,2011automatic,wang2021self} require bounding boxes of each target object as the ground truth, making annotation laborious and time-consuming. In counting scenarios with only point locations as ground truth, they struggle to introduce pseudo-labeled boxes for learning. Additionally, detection-based methods are not robust enough for occlusion and scale variations, making them less suitable for large-scale counting tasks. Despite these challenges, there are ongoing research efforts that explore the detection-based paradigm as a potential solution to the counting problem.
LST-CNN \cite{sam2020locate} and Crowd-SDNet \cite{wang2021self} employ the generation of pseudo-labeled boxes based on ground truth point labels. This enables the model to not only perform crowd counting but also predict the centroids and sizes of individuals.
Additionally, Crowd-SDNet continuously refines the pseudo-labeled boxes during training, thereby improving the counting accuracy and the quality of object position and size predictions. However, it should be noted that, despite these efforts, the pseudo-labeled bounding boxes are still subject to inaccuracies, which can have a negative impact on the final counting results.

The regression methods \cite{MAN,LiuCWCLL20MM,asnet,MCNN,ADCrowdNet19cvpr,CAN,YuanQLWCCL20Neuro} commonly exploit density estimation, which is most extensively studied and generally yields superior performance compared to other schemes.
Previous studies within this framework have provided innovative solutions to address more challenging aspects such as occlusions and large variations of scale and density.
MCNN \cite{MCNN} and ADCrowdNet \cite{ADCrowdNet19cvpr} address the issue of scale variations by employing multi-column architectures to extract features at different scales. CAN \cite{CAN} and PGCNet \cite{PGCNet} leverage perspective maps to generate more accurate density maps for object regions with different scales. 
To handle the challenge of density variation, AS-Net \cite{asnet} trains a dedicated model to classify people in the image into different density levels and employs separate network branches for different density regions. 
Liu et al. \cite{LiuCWLLL21cvpr} learns a cross-modal collaborative representation for RGBT crowd counting.
Sun et al. \cite{boost} replace traditional convolutional neural networks with Transformers to optimize feature representations and improve performance.
CCTrans \cite{cctrans} employs a Transformer-based model Twins \cite{chu2021twinsnips} as its backbone, integrating a feature pyramid module to fuse high-level semantics and low-level features for crowd counting.
Zhao and Li \cite{DDMD} introduce deformable convolutions to fit the Gaussian kernel variation and enhance the model's adaptability to scale changes caused by perspective effects. 
CrowdCLIP \cite{crowdclip} exploits the image-text alignment capability of CLIP \cite{CLIP} to achieve unsupervised learning of the model through patch-level image-text matching for crowd counting.
So far, density map-based approaches have generally exhibited superior counting accuracy compared to other methods, but their generated results often suffer from blurriness, limiting their usefulness in more complex downstream tasks that require precise object localization.

Recently, several localization-based methods \cite{p2pnet,TopoCount,FIDTM} have emerged to address crowd counting. These methods not only count the number of objects but also provide specific object locations, striking a compromise between traditional detection-based and density map-based approaches.
TopoCount \cite{TopoCount} introduces a novel topological approach that treats the counting task as the prediction of a binary mask, referred to as a topological map. A one-to-one relationship is established for each component within the topological map and a target point. 
Liang et al.\cite{FIDTM} propose the Focal Inverse Distance Transform Map as a novel representation for density maps. Unlike previous density map schemes, this method separates each point in the density map, which helps to indicate the object locations.
P2PNet \cite{p2pnet} directly predicts the object locations by employing a one-to-one matching strategy between the predicted points and the ground truth. 
Although the current leading localization-based methods may not surpass density map-based methods in terms of counting accuracy, they demonstrate promising potential and offer distinct advantages of object locations in addressing complex downstream tasks.

\subsection{Class-Agnostic Counting}

In contrast to class-specific counting, class-agnostic counting (CAC) represents a more generalized object counting task that aims to count the objects of an arbitrary class with only a few exemplars provided at test time.  
It offers the advantage of applicability across various scenarios without the necessity of model retraining. 
Meanwhile, this task becomes significantly more challenging as the model is required to count objects from previously unseen classes during testing.
Recently, some works \cite{gmn,famnet,bmnet,SAFECount,ZhaiDQWZFC24kbs} have been presented to address this newly emerging task.
Lu et al. \cite{gmn} propose a generic matching network for class-agnostic counting, where the exemplar features are upsampled and concatenated with the image query features, and then the similarity map is finally obtained for counting.
To address the lack of adequate datasets, Ranjan et al. \cite{famnet} introduce a benchmark FSC-147 for the CAC task and propose FamNet. FamNet \cite{famnet} uses multiple correlation maps between exemplars and image features for density prediction and fine-tunes the model parameters using an adaptive loss at test time. 
In addition, they employ few-shot object detection methods FR \cite{fr} and FSOD \cite{fsod} to tackle this task for comparison.
However, due to the differences in task objectives, the feature requirements and architecture designs of more generalized detection methods are tailored to effectively detect multiple classes of objects. This often comes at the cost of partially sacrificing the ability to detect very small objects, rendering them less suitable for practical counting scenarios. Substantial efforts are required to modify these methods to be more suitable for counting.
Inspired by the evident advantages of density estimation observed in class-specific counting, most leading CAC methods adopt a density regression scheme.

\begin{figure*}[!t]
    \centering
    \includegraphics[width=0.935\textwidth]{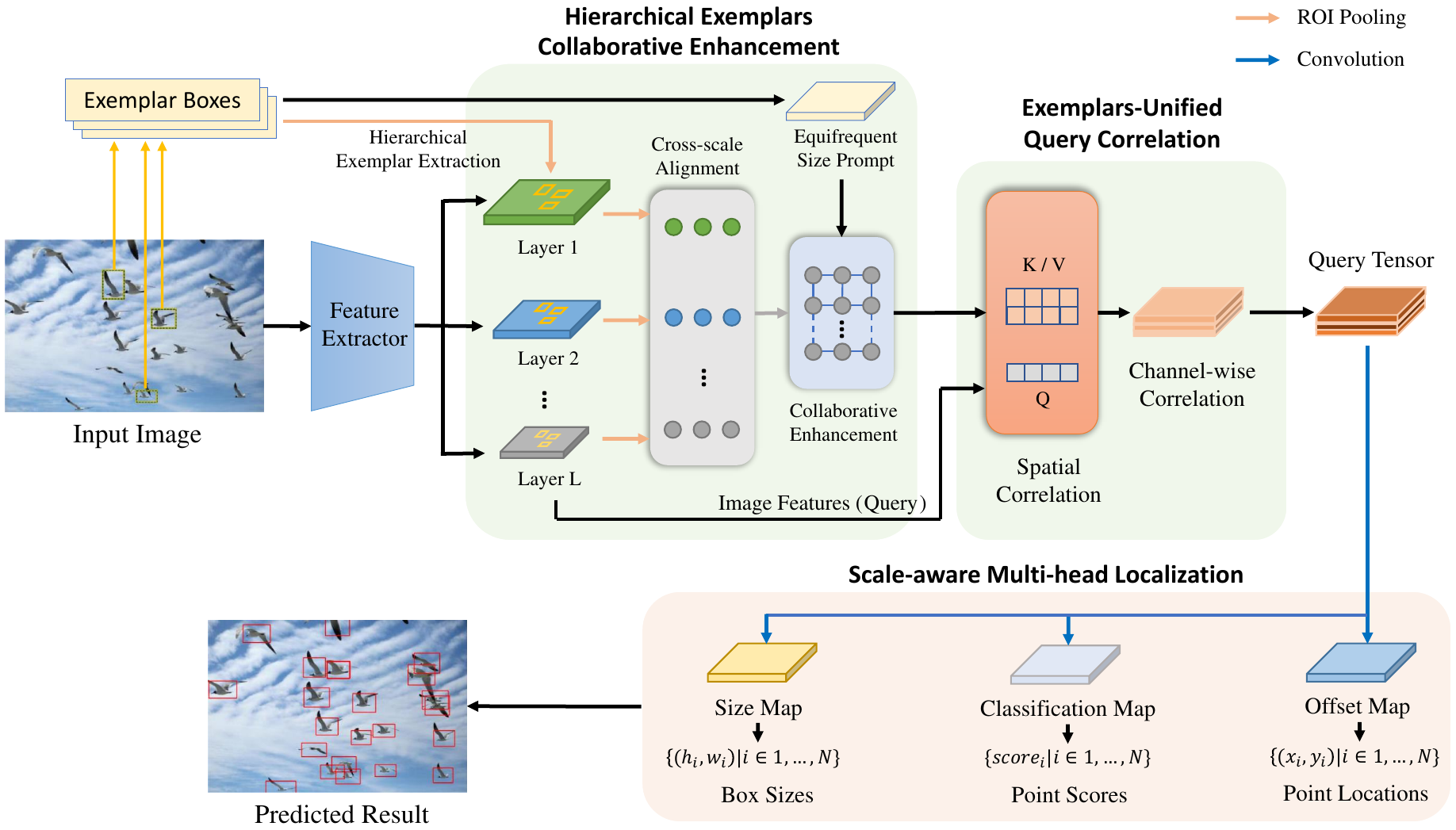}
    \caption{Illustration of the proposed SQLNet framework. It mainly consists of three modules to accomplish the query and localization stages for class-agnostic counting. In the query stage, by multi-scale feature mining and equifrequent size prompt, the Hierarchical Exemplars Collaborative Enhancement (HECE) module produces rich discriminative representations of the target class from limited exemplars. They are fed into the Exemplars-Unified Query Correlation (EUQC) module to interact with the query image feature in a unified manner to obtain the correlated query tensor.
    In the localization stage, the Scale-aware Multi-head Localization (SAML) module predicts the confidence, location, and size of each potential object, and a scale-aware localization loss is specially introduced for learning. The modules responsible for the query and localization stages are distinguished by light green and light orange backgrounds.}
    \label{fig:pipeline}
\end{figure*}

Gong et al. \cite{rcac_eccv} conduct a comprehensive analysis of FamNet \cite{famnet} and propose the use of exemplar feature augmentation and edge matching to enhance the model's robustness against intra-class diversity.
Shi et al. \cite{bmnet} propose a similarity-aware framework BMNet+ that jointly learns representation and similarity metric for density estimation.
Lin et al. \cite{SPDCN_bmvc} introduce a scale-prior deformable convolution that integrates the scale information of the exemplars into the counting network backbone, thereby enhancing the robustness of exemplar-related feature extraction.
You et al. \cite{SAFECount} present an iterative framework, SAFECount, that progressively refines the exemplar-related features based on the correlation between the image and exemplars for density map regression.
Wang et al. \cite{WangYWLC24nn} introduce a scale-aware transformer-based feature fusion module to enhance density map regression.
CAC is simplified by \cite{WangX0024aaai} in an extract-and-match manner, where a vision transformer (ViT) is used for feature extraction and similarity matching  simultaneously.  Huang et al. \cite{HuangD0ZS24cvpr} present a generalized framework for both few-shot and zero-shot object counting, where the powerful Segment Anything Model (SAM) \cite{KirillovMRMRGXW23iccv} is used to produce segmentation mask proposals of objects. Different from the CAC task, Dai et al. \cite{DaiLC2024CVPR} propose a new Referring Expression
Counting (REC) task that aims to count objects with different attributes within the same class using text descriptions. They combine the vision-language model CLIP \cite{CLIP} with global-local feature fusion to count the objects with the given attributes. 

Different from the above methods, this work investigates a localization-based CAC method that can achieve state-of-the-art counting performance. 
We introduce a scale-aware localization learning scheme that takes full advantage of object locations and exemplar scales for supervision, enabling our model to predict the location and size of each object to facilitate accurate counting.
Moreover, we design novel architectures to acquire rich discriminative representations of the target class and conduct their interaction with the image features through an exemplars-unified manner, thus capturing sufficient correlation information in the query stage for prediction.

\section{Method}
\label{sec:method}

This section first presents the problem formulation for the CAC task and then provides an overview of the proposed SQLNet, followed by a detailed description of each module.

\subsection{Problem Formulation}\label{sec:Problem}

Given an input image where several exemplars of an arbitrary class are provided, the goal of class-agnostic counting (CAC) is to count all the objects of the target class in the input image.
Formally, let $I$ denote the input image, and the few exemplars are specified with bounding boxes, denoted by $\mathbf{B} = {\{B_{i} \}_{i=1}^{N^B}}$. $N^B$ is the number of exemplars.
Each object of the specified class in the input image is annotated with a point as the ground truth in a CAC dataset. Let $\mathbf{D} = {\{ d_{i}=(x_{i}, y_{i} )\}_{i=1}^{M}}$ denote the annotated objects, where $(x_{i}, y_{i} )$ is the coordinates of the $i$-th object in the image and $M$ is the number of ground truth objects.
To ensure the generality of a model, the classes in the training, validation, and test set of a CAC dataset have no intersection.

\subsection{Architecture Overview}

Figure \ref{fig:pipeline} illustrates the architecture of the proposed SQLNet. 
In contrast to previous leading methods that count the number via density map regression, our method works beyond counting by accurately locating each object with a bounding box. 

Our SQLNet model mainly consists of three modules, i.e., Hierarchical Exemplars Collaborative Enhancement (HECE), Exemplars-Unified Query Correlation (EUQC), and Scale-aware Multi-head Localization (SAML). 
The first two modules complete the query stage of our model, and the last one is employed for the localization stage. The HECE module learns rich discriminative representations of the target class from the few provided exemplars by exploring multi-scale exemplars collaborative interactions with equifrequent size prompt. Then the enhanced object representations, together with the feature of the query image, are fed into the EUQC module, where spatial and channel-wise correlations are conducted in an exemplar-unified scheme. Afterwards, the EUQC module outputs a query tensor, where each location indicates its correlation with all exemplars. Finally, the SAML module predicts each potential object instance's confidence, location, and size with three heads stemming from the query tensor. Though the multi-head prediction in the SAML module is in a similar form to conventional object detection, the learning paradigm is quite different. Our method exploits flexible point associations to provide location supervision and uses the sizes of exemplars to provide scale supervision in learning. 
In the following subsections, we will describe each module in detail.

\subsection{Hierarchical Exemplars Collaborative Enhancement} \label{sec:Enhancement}

Since only a few (typically around three, even one) exemplars of an arbitrary class are provided in the input image, the HECE module is designed to learn rich discriminative representations of this specified class from exemplars at different scales via a collaborative enhancement mechanism. 

\vspace{0.5ex}\noindent\textbf{Hierarchical Exemplar Extraction.}~
We employ a well-studied network structure (e.g., ResNet \cite{resnet}) as the backbone of our model, denoted as the feature extractor. Given an input image, the feature extractor will produce hierarchical and gradually abstract feature representations of the image, where the features of lower layers reflect more detailed information about the image while the features of higher layers reflect more abstract semantic information. 
We use the output of $L$ layers from the feature extractor to obtain feature representation at $L$ different scales. The bounding box of each exemplar is projected onto each scale to find its corresponding representation.

\vspace{0.5ex}\noindent\textbf{Cross-Scale Feature Alignment.}~
The representations of an exemplar from different scales generally have different feature dimensions. We introduce cross-scale feature alignment to make them have the same dimension for subsequent collaborative enhancement. Specifically, let $F_{i,j} \in \mathbb{R}^{H_{i,j}\times W_{i,j} \times C_{j}}$ denote the representation of exemplar $i$ at scale $j$, where $H_{i,j}$ and $W_{i,j}$ denote the width and height of exemplar $i$ at scale $j$, and $C_{j}$ is the channel number of feature at scale $j$. We apply region-of-interest (ROI) pooling to $F_{i,j}$ and obtain a $C_{j}$-dimensional feature vector $f_{i,j} \in \mathbb{R}^{C_{j}}$. 
To ensure that the feature vectors of exemplar $i$ at all scales have the same dimension, we use a linear projection function to perform cross-scale feature alignment. Concretely, we use the channel number $C_{L}$ at scale $L$ as the standard. For scale $j$, if its channel number is not equal to $C_{L}$, i.e., $C_j \neq C_L$, the feature vector $f_{i,j}$ of exemplar $i$ at scale $j$ will be mapped to a $C_{L}$-dimensional feature vector $\hat{f}_{i,j} \in \mathbb{R}^{C_{L}}$, formulated as:
\begin{equation}\label{eq:}
\hat{f}_{i,j}=\phi(f_{i,j}),
\end{equation}
where the linear projection function is implemented as a neural network with one fully-connected layer.

\begin{figure}[!t]
    \centering
    \includegraphics[width=0.9\linewidth]{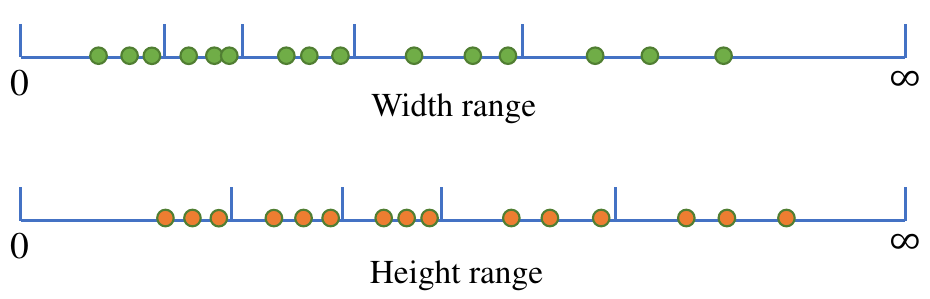}
    \caption{Illustration of equifrequent division for size prompt. Each point denotes the width or height of an exemplar. The range of with and height of exemplars are divided into $T$ intervals so that each interval has roughly the same number of exemplars.}
    \label{fig:equifreq}
\end{figure}

\vspace{0.5ex}\noindent\textbf{Equifrequent Size Prompt.}~
\label{sec:ESP}
In the above procedure, the exemplar representation loses the size information, since the feature is extracted using ROI pooling, which compresses the spatial dimension. However, it is important to make the model fully perceive the scale information of the few exemplars. To resolve this, we incorporate size prompt into the exemplar representation before collaborative enhancement.
Specifically, we design non-shared size prompts for width and height, respectively, since it is crucial to distinguish different objects that vary in width and height.
Moreover, to provide a fixed number of learnable size prompt embeddings that are robust to small variations, we design an equifrequent size prompt scheme that exemplars falling within the same size range in terms of width and height will share the same size prompt, as illustrated in Figure \ref{fig:equifreq}. Let's take the width as an example to explain. We acquire the width values of all exemplars annotated in the training set and divide the width range into $T$ different intervals so that each interval has roughly the number of exemplars. Assume the total number of exemplars is $N_{a}$ and the list of the width values of all exemplars is $\{w_1,w_2,...,w_{N_a}\}$, which are sorted in ascending order. Then the $k$-th interval can be calculated as:
\begin{equation}\label{eq:interval}
\mathcal{U}^w_{k} = \left(w_{(k-1)\cdot\lfloor\frac{N_a}{T}\rfloor},\, w_{k\cdot\lfloor\frac{N_a}{T}\rfloor}\right), k=1,...,T-1
\end{equation}
where the operator $\lfloor\cdot\rfloor$ indicates the largest integer not larger than the given number.
The upper bound of the last interval (i.e., $\mathcal{U}^w_{T}$) is infinity. The height range is divided similarly. Finally, we obtain $T$ intervals for width and height, respectively, which are corresponding to $2T$ learnable size prompt embeddings $\{E^{w}_k\}_{k=1}^{T}$ and $\{E^{h}_k\}_{k=1}^{T}$.

For the $i$-th exemplar in the input image, its size prompt is obtained by concatenating the prompts corresponding to its width and height. Specifically, $B_i$ is the bounding box of exemplar $i$, and we let $B_i^w$ and $B_i^h$ denote its width and height. The size prompt embedding $E_i^{s}$ of exemplar $i$ is formulated as:

\vspace{-2.5ex}\begin{gather}\label{eq:size_embedding}
E_i^{s} = [E^{w}_{a},\,E^{h}_{b}] \\
a=\varphi_w(B_i^w),\; b=\varphi_h(B_i^h)
\end{gather}
where the operators $\varphi_w(\cdot)$ and $\varphi_h(\cdot)$ map the width and height to the corresponding intervals, respectively.

\begin{figure*}[!t]
    \centering
    \includegraphics[width=0.98\textwidth]{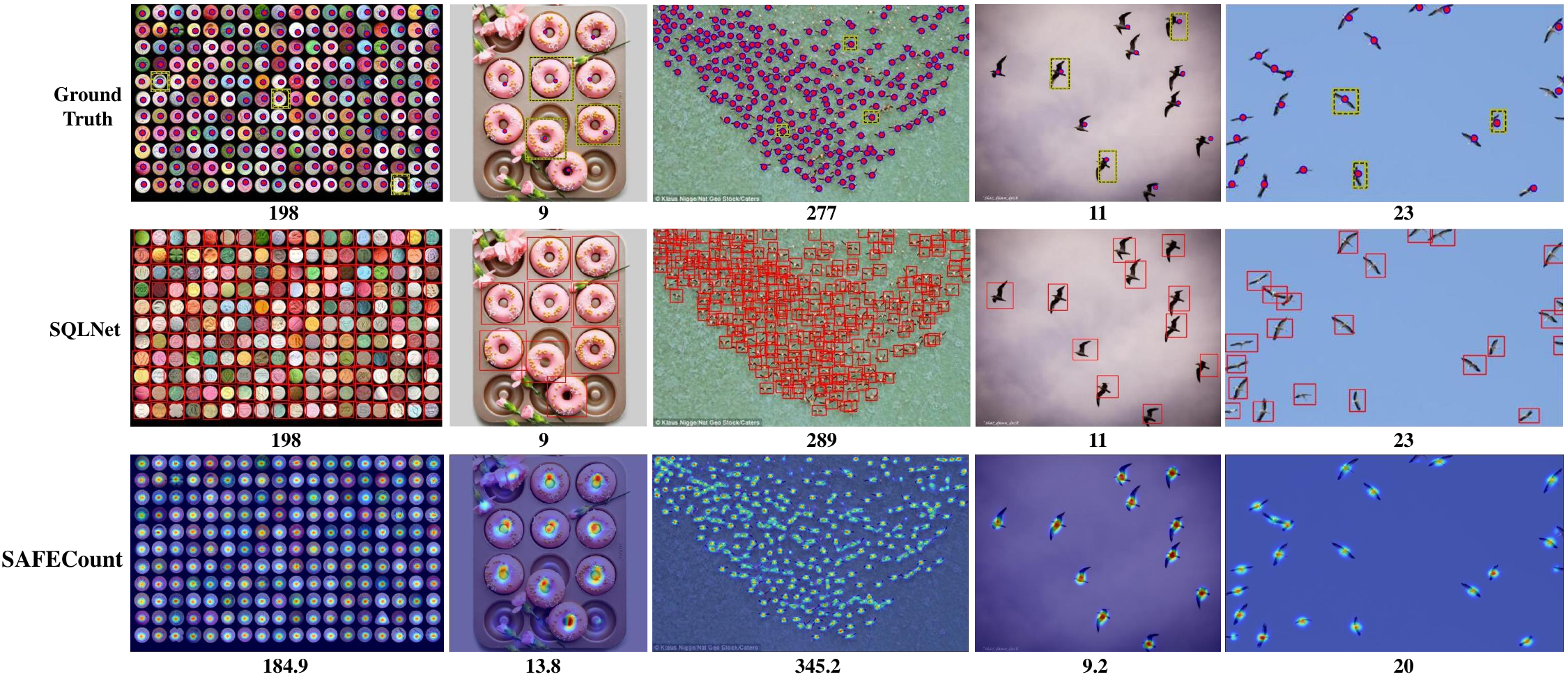}
    \caption{Qualitative comparison results on FSC-147. The first row shows the ground truth in the query images, where the purple points indicate the locations of object instances and the yellow boxes denote the provided exemplars. As shown in the second row, our SQLNet achieves counting by predicting each object's location and size. In the third row, the predicted density maps by SAFECount are visualized on the query images. The numbers below the images indicate the counting results.}
    \label{fig:compare}
\end{figure*}

\vspace{0.5ex}\noindent\textbf{Exemplars Collaborative Enhancement.}
\label{para:ExCA}
Since different objects of the same class can vary greatly in appearance on the same image, it is rather difficult to capture the discriminative commonalities of the same class from only a few exemplars. 
Therefore, rather than handle the exemplar representations separately, we propose to strengthen their discriminative commonalities via collaborative enhancement. 
Inspired by the learning power of Transformer \cite{AshishVaswani2017AttentionIA,Wu2024PersonSearch,Pu2024Transformer}, we formulate the collaborative enhancement mechanism in a similar form to the self-attention mechanism used in Transformer. 

Specifically, we have $N^B$ exemplars in the input image and extract multi-scale representations from $L$ layers, and thus we obtain $N_{e}=N^B\times L$ exemplar presentations in total. By treating each representation as a token, we have the following token list:
\begin{equation}
	\mathbf{x} =\left[ \mathcal{F}^i + E^i_{s}\;|\;i=1,...,N_{e}\right]
\end{equation}
where $\mathcal{F}^i$ denotes the feature presentation and $E^i_{s}$ is the size prompt embedding. The collaborative enhancement is formulated as:

\vspace{-2.5ex}\begin{gather}\label{eq:Attention}
Attention(Q,~K,~V) = softmax \left(\frac{QK^{\top}}{\sqrt{d_e}}\right)V\\
Q=W_1^{Q}\mathbf{x},~K=W_1^{K}\mathbf{x},~V=W_1^{V}\mathbf{x}
\end{gather}
where $W_1^{Q}$, $W_1^{K}$ and $W_1^{V}$ are learnable parameters, and $d_e$ is the dimension of feature representation.
In this formulation, each exemplar representation fully interacts with all exemplar representations in the list through the multiplication of $Q$ and $K$ and then incorporates the correlated information into itself by the multiplication of $V$. In this way, each exemplar representation can effectively enhance its discriminative parts of the target class. In addition, it can be efficiently computed via parallel processing.

The collaborative enhancement is achieved with $L_{c}$ Transformer layers in our implementation, i.e.,

\vspace{-2.5ex}\begin{align}
	\mathbf{z}_0 &=\mathbf{x}, \\
        \mathbf{z}'_l& = MHSA(LN(\mathbf{z}_{l-1}))+\mathbf{z}_{l-1}, \ \ &l=1,...,L_{c} \\
	\mathbf{z}_l& = MLP(LN(\mathbf{z}'_{l}))+\mathbf{z}'_{l}, \ \  &l=1,...,L_{c}
\end{align}
where $MHSA(\cdot)$ is the multi-head self-attention operation in Transformer, $LN(\cdot)$ denotes the layer normalization, and $MLP(\cdot)$ is a multi-layer perceptron. 

Upon computation, we will obtain a list of enhanced exemplar representations $\mathbf{\hat{x}}=\mathbf{z}_{L_c}$, whose number is also $N_e$.
They serve as a collection of rich discriminative representations of the target class, which are used to discover the potential objects of the same class in the input image afterward.

\subsection{Exemplars-Unified Query Correlation}
\label{sec:EUQC}
The EUQC module conducts interactions between the enhanced representations of the target class and the feature of the input image to output their correlation information.  
Different from previous works that perform the interaction in an exemplar-by-exemplar manner and then aggregate the output, we propose to perform the exemplar-image interaction in an exemplars-unified way. Specifically, given the feature of the input image as the query, spatial and channel-wise correlations are successively carried out by the EUQC module to output a query tensor, where each location indicates its correlation information with all exemplars.

\vspace{0.5ex}\noindent\textbf{Spatial Correlation.}
To perform the spatial correlation of the input image and the enhanced class representations, we explore a revised formulation from the above collaborative enhancement, which meets our purpose and brings great benefits. 
The reasons are twofold. First, it enables taking all the exemplar representations as a whole to participate in the interaction. Second, as aforementioned, such formulation can achieve full interaction and be performed with efficient computing. 

Specifically, we can use Equation (\ref{eq:Attention}) to formulate the spatial correlation, but the $Q$, $K$, and $V$ are computed differently to represent the features of the input image and the class representations. In the new formulation, we take the image feature $F^I_L$ output from the $L$-th layer of the feature extractor as the query. Its channel number is equal to the dimension of the exemplar representation, which derives from the cross-scale feature alignment in Section \ref{sec:Enhancement}. Let $W_L$, $H_L$, and $C_L$ denote the width, height, and channel number of $F^I_L$, respectively. We take the feature vector at each spatial position of $F^I_L$ as a token, and obtain the following token list:
\begin{equation}
	\mathbf{q} =\left[ \mathcal{F}_q^i + E^i_{pos}\;|\;i=1,...,N_{q}\right]
\end{equation}
where $\mathcal{F}_q^i$ denotes the feature vector at the $i$-th position and $E^i_{pos}$ is the corresponding position embedding that adopts sinusoidal assignment \cite{AshishVaswani2017AttentionIA}. $N_{q}$ denotes the number of tokens and $N_{q}=W_L\times H_L$.

To conduct spatial correlation in an exemplar-unified way, we compute $Q$, $K$, and $V$ as follows:
\begin{equation}\label{eq:spatial}
Q=W_2^{Q}\mathbf{q},~K=W_2^{K}\mathbf{\hat{x}},~V=W_2^{V}\mathbf{\hat{x}}
\end{equation}
where $\mathbf{q}$ and $\mathbf{\hat{x}}$ are the token lists that represent the image feature and the enhanced exemplar representations. $W_2^{Q}$, $W_2^{K}$ and $W_2^{V}$ are learnable parameters. By substituting them into Equation (\ref{eq:Attention}), we make the image features fully interact with all exemplar representations and obtain correlation information for each spatial location. Similar to the collaborative enhancement, we can use $L_q$ Transformer layers for the implementation of the spatial correlation, but special care should be paid to the detailed design so that the calculation is consistent with Equation (\ref{eq:spatial}).

\vspace{0.5ex}\noindent\textbf{Channel-wise Correlation.} 
In the output $\mathcal{O}_q$ of spatial correlation, channel-wise information of image features and the class representations are implicitly fused. However, considering that different channels of the correlation information are not equally important for later prediction, we propose the channel-wise correlation that explicitly models the channel-wise interdependencies of the exemplar representations and adaptively recalibrates the importance of different channels of the correlation information.
Specifically, we exploit a network design similar to the squeeze and excitation network \cite{senet} for this purpose. But different from \cite{senet} that operates on the input feature itself, we apply the operation between $\mathcal{O}_q$ and the exemplar representations $\mathbf{\hat{x}}$. 

Concretely, all the exemplar representations $\mathbf{\hat{x}}$ are mapped to a $d_e$-dimensional feature vector $G_e$ by global average pooling. Then, it is mapped to a weighting vector $G_w \in \mathbb{R}^{d_e}$ whose elements denote the recalibration weights for the corresponding channels of the correlation information $\mathcal{O}_q$, formulated as: 
\begin{equation}\label{eq:channel}
G_w=\psi(G_e),
\end{equation}
where the function $\psi(\cdot)$ first reduces the dimension of $G_e$ and then increases the dimension back to learn a nonlinear interaction between channels, which is well validated by previous work \cite{senet}.
In architecture design, $\psi(\cdot)$ can be implemented as a small network with two fully-connected layers (one for dimension reduction and one for dimension increase), followed by a softmax layer.

After the weighting vector $G_w$ is obtained in an exemplars-unified way, the final query tensor is calculated as:
\begin{equation}\label{eq:tensor}
\mathcal{O}'_q=\mathcal{O}_q \odot G_w ,
\end{equation}
where the operator $\odot$ indicates channel-wise multiplication for importance recalibration. It multiplies each weight in $G_w$ to the corresponding channel of $\mathcal{O}_q$.

\begin{table*}[!t]
\centering
\small
\caption{Evaluation results on FSC-147 with state-of-the-art methods. The best results are highlighted in bold.}
\label{table:result}
\begin{tabular}{ccccccc}
\hline
\multirow{2}{*}{Method} & \multirow{2}{*}{Year} & \multirow{2}{*}{Paradigm} & \multicolumn{2}{c}{Val} & \multicolumn{2}{c}{Test} \\ \cline{4-7} 
 &  &  & MAE & RMSE & MAE & RMSE \\ \hline
MAML \cite{maml} & 2017 & Detection & 25.54 & 79.44 & 24.9 & 112.68 \\
GMN \cite{gmn} & 2018 & Detection & 29.66 & 89.81 & 26.52 & 124.57 \\
FR \cite{fr} & 2019 & Detection & 45.45 & 112.53 & 41.64 & 141.04 \\
FSOD \cite{fsod} & 2020 & Detection & 36.36 & 115.00 & 32.53 & 140.65 \\  \hline
FamNet \cite{famnet} & 2021 & Regression & 23.75 & 69.07 & 22.08 & 99.54 \\
RCAC \cite{rcac_eccv} & 2022 & Regression & 20.54 & 60.78 & 20.21 & 81.86 \\
BMNet+ \cite{bmnet} & 2022 & Regression & 15.74 & 58.53 & 14.62 & 91.83 \\
SPDCN \cite{SPDCN_bmvc} & 2022 & Regression & 14.59 & 49.97 & 13.51 & 96.80 \\
SAFECount \cite{SAFECount} & 2023 & Regression & 15.28 & 47.20 & 14.32 & 85.54 \\ \hline
PseCo \cite{HuangD0ZS24cvpr} & 2024 & Segmentation & 15.31 & 68.34 & 13.05 & 112.86 \\ \hline
SQLNet(Ours) & - & Localization & \textbf{12.40} & \textbf{42.30} & \textbf{12.49} & \textbf{80.85} \\ \hline
\end{tabular}
\end{table*}

\subsection{Scale-Aware Multi-head Localization}

The Scale-Aware Multi-head Localization (SAML) module in the localization stage aims to locate each potential object instance of the target class. Besides the position, our method is aware of the object scale, i.e., also predicting the size of each object. As stated earlier, except for the given exemplars, the ground truth objects are annotated with points rather than bounding boxes. Therefore, it is quite challenging to predict the sizes of objects together. 

Specifically, our SAML module predicts each instance's confidence, location, and size with three heads stemming from the query tensor $\mathcal{O}'_q$ output by the EUQC module. 
Each head is a branch of a convolutional neural network. In our implementation, the convolution architectures of the three branches are kept the same for simplicity, which consists of three convolutional layers interleaved with ReLU activations. 
In the form of network design, it is similar to conventional object detection methods that use multiple heads for prediction, but the learning paradigm is different, which will be detailed afterward.

\begin{figure}[!t]
    \centering
    \includegraphics[width=0.8\linewidth]{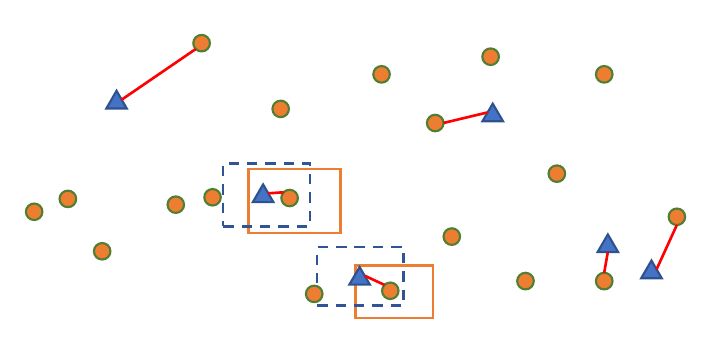}
    \caption{Illustration of the scale-aware localization loss. The blue triangles denote the ground truth object positions, and the yellow circles denote the point proposals. The dashed-line boxes denote the provided exemplars. In learning, each ground truth point is dynamically matched to a point proposal via the Hungarian algorithm, in contrast to conventional object detection that a ground truth object is matched to a fixed anchor proposal. In each iteration, all unmatched point proposals are considered negative samples. If a point proposal is matched to a given exemplar, its predicted size (denoted by the solid-line box) will be compared with the size of the exemplar for scale regularization in the loss. }
    \label{fig:loss}
\end{figure}

Assume the width and height of the query tensor $\mathcal{O}'_q$ are $W_q$ and $H_q$, respectively.
Each pixel on $\mathcal{F}_r$ roughly corresponds to a patch size of $s\times{s}$ in the original image. For each patch $i$, we define a fixed set of anchor points $A^i=\{A^i_{j}=(x^i_{j}, y^i_{j})~|~j\in \{1,...,N_a\}\}$, which are uniformly distributed on the patch. An object proposal will be generated for each anchor point, so there will be a total of $N_a\times W_q \times H_q$ object proposals. We ensure the object proposals are overpopulated, i.e., the number of proposals is much more than that of the ground truth objects.

The three heads of the SAML module output an offset map, a classification map, and a size map, respectively. 
With the predicted offset map, a set of point proposals $P$ is generated from the anchor points. To be specific, let $( \Delta x^i_{j}, \Delta y^i_{j})$ denote the predicted offset for the anchor point $A^i_{j}$, the coordinates of the corresponding point proposal  $\hat{A}^i_{j}=(\hat{x}^i_{j}, \hat{y}^i_{j}) \in P$ is obtained by:
\begin{equation}\label{eq:offset}
	\hat{x}^i_{j} = x^i_{j}+\alpha \Delta x^i_{j}, \quad
	\hat{y}^i_{j} = y^i_{j}+\alpha \Delta y^i_{j} ,
\end{equation}
where $\alpha$ is a scaling parameter. 

The classification map is a set of point scores $\xi$, where the $j$-th point score $\xi_j$ denotes the confidence of the $j$-th point proposal belonging to the target class. For each point proposal $\hat{A}^i_{j}$, we also predict its size $(w^i_{j}, h^i_{j})$, which is contained in the size map $S$. The final size of a proposal is calculated as follows:
\begin{equation} \label{eq:size}
	\hat{w}^i_{j} = \beta w^i_{j}, \quad  \hat{h}^i_{j} = \beta h^i_{j} ,
\end{equation}
where $\beta$ is also a scaling parameter.

For inference, the SAML module generates all the point proposals with the corresponding point scores and sizes. Each point proposal with a score larger than a threshold (commonly set as 0.5) is considered as a target object.

\vspace{0.5ex}\noindent\textbf{Learning.}~
We present the learning paradigm and the optimization loss in this section since it is closely related to the SAML module. We term the loss in our learning paradigm as the scale-aware localization (SAL) loss (illustrated in Figure \ref{fig:loss}), which well leverages flexible location associations and the sizes of exemplars to provide supervision.

To the best of our knowledge, we are the first to introduce a scale-aware localization-based scheme to address the CAC task. Inspired by \cite{detr,p2pnet}, we use the Hungarian algorithm to find the best matching between the ground truth object positions and the point proposals. Concretely, given the ground truth points $G=\left\{G_i~|~i=1,...,M\right\}$ and the point proposals  $P=\left\{P_j~|~j=1,...,N\right\}$, our goal is to match each ground truth point to a point proposal so that the matching cost is minimal. The match cost between a ground truth point $G_i$ and a point proposal $P_j$ is defined by considering the point score $\xi_j$ and the distance, i.e.,
\begin{equation} \label{eq:size_D}
	D(G_i,P_j) = -\xi_j+\eta \|G_i-P_j\|_2 , 
\end{equation}
where $\|\cdot\|_2$ denotes the $L_2$ norm (i.e., the Euclidean distance), and $\eta$ is a balancing parameter. Therefore, the matching cost to be minimized is as follows:
\begin{equation} \label{eq:size_match_loss}
	\mathcal{D}=\sum_{i=1}^M D(G_i,P_{\zeta(i)}),
\end{equation}
where $\zeta(i)$ denotes the index of point proposal that is matched to the $i$-th ground truth point $G_i$. This formulation can be solved by the Hungarian algorithm efficiently.

After the best match is obtained, all the unmatched point proposals are considered as negative samples. Furthermore, if a point proposal is matched to a given exemplar, its predicted size will be compared with the size of the exemplar for scale regularization. Therefore, the final loss for model learning is defined as:
\begin{equation}
\mathcal{L} = \mathcal{L}_{cls} + \lambda_1 \mathcal{L}_{loc} + \lambda_2 \mathcal{L}_{size}  ,
\end{equation}%
where $\lambda_{1}$ and $\lambda_{2}$ are weight factors for balancing the effect of location and size supervision, and $\mathcal{L}_{cls}$, $\mathcal{L}_{loc}$ and $\mathcal{L}_{size}$ are the classification, location, and size losses, respectively. Specifically, We use cross-entropy to optimize the point score for the classification loss $\mathcal{L}_{loc}$, adopt the Euclidean distance for the location loss $\mathcal{L}_{loc}$, and employ the Manhattan distance to evaluate the size loss $\mathcal{L}_{size}$. They are formulated as follows:

\vspace{-2ex}\begin{align}
\label{eq:Loss}
    \mathcal{L}_{cls} &=-\frac{1}{N} \sum^{N}_{j=1}\big\{ \mathbbm{1}_j log\ \xi_{j} + \gamma (1-\mathbbm{1}_j)\ log(1- \xi_{j})\big\}\\
    \mathcal{L}_{loc} &= \frac{1}{M}\sum^M_{i=1}\left(||P_{\zeta(i)}^x-G_i^x||^2_2 + ||P_{\zeta(i)}^y-G_i^y||^2_2\right)\\
    \mathcal{L}_{size} &= \frac{1}{N^{B}}\sum^{N^{B}}_{k=1} \left(||S_{\zeta(k)}^w-B_{k}^w||_1 + ||S_{\zeta(k)}^h-B_{k}^h||_1 \right)
\end{align}%
where $\mathbbm{1}_j$ is an indicator that takes 1 if the $j$-th point proposal is matched to a ground truth point and takes 0 otherwise, and $\gamma$ is the weighting factor for negative proposals. 
$(G_i^x, G_i^y)$ and $(P_{\zeta(i)}^x, P_{\zeta(i)}^y)$ are the coordinates of the $i$-th ground truth point and the point proposal matched to it.
$S_{\zeta(k)}^w$ and $S_{\zeta(k)}^h$ are the predicted size (width and height) of the point proposal that is matched to the $k$-th exemplar, and $B_k^w$ and $B_k^h$ are the width and height of the $k$-th exemplar. The operator $\|\cdot\|_1$ denotes the $L_1$ norm.

\begin{table*}[!t]
\centering
\small
\caption{Evaluation results on Val-COCO and Test-COCO. The best results are highlighted in bold. $^{\dag}$ denotes the results obtained based on the official code and model.
}
\label{table:coco}
\begin{tabular}{ccccccccc}
\hline
\multirow{2}{*}{Method}  & \multirow{2}{*}{Paradigm} & \multicolumn{3}{c}{Val-COCO} & \multicolumn{3}{c}{Test-COCO} \\ \cline{3-8} 
 &  &  MAE & RMSE & nAP(\%) & MAE & RMSE & nAP(\%)\\ \hline
Faster-RCNN \cite{fastrcnn}  & Detection  & {52.79} & 172.46 & 8.61 & {36.20} & 79.59 & 37.91\\ 
RetinaNet \cite{retinanet}  & Detection  & {63.57} & 174.36 & 6.50 & {52.67} & 85.86 & 44.17\\ 
Mask-RCNN \cite{maskrcnn}  & Detection  & {52.51} & 172.21 & 9.72 & {35.56} & 80.00 & 34.85\\  \hline
FamNet \cite{famnet} & Regression  & {39.82} & 108.13 & - & {22.76} & 45.92 & - \\ 
BMNet+$^{\dag}$  \cite{bmnet}  & Regression & {26.55} & 93.63 & - & {12.38} & 24.76 & - \\ 
SAFECount \cite{SAFECount}  & Regression & {22.85} & {63.33} & - & {13.13} & \textbf{23.68} & - \\ \hline
SQLNet(Ours)  & Localization  & {{\textbf{21.21}}} & \textbf{61.14} & \textbf{60.42} & {{\textbf{11.04}}} & {24.38} & \textbf{78.11} \\ \hline
\end{tabular}\vspace{1ex}
\end{table*}

As aforementioned, state-of-the-art CAC methods all adopt a density map regression paradigm, which sums up the density map to obtain the number of objects belonging to the target class. In contrast, by exploiting the scale-aware localization paradigm, we achieve a superior class-agnostic counting performance by accurately locating each object of the target class as well as predicting its size.
To our best knowledge, we are the first to explore such a localization-based paradigm that can outperform state-of-the-art CAC methods. 

We further clarify the characteristics of our scale-aware localization learning paradigm by comparing it with some closely related methods that are not specified for the CAC task.  
Recent few-shot object detection methods \cite{fr,fsod} also predict the bounding box of objects with few exemplars of unseen classes. However, their schemes are different from ours. They generally model the relationship between the object proposals and the exemplars at the back-end of the network, e.g., by feature re-weighting or matching, to fulfill object detection. Moreover, in their task, each object is annotated with a ground truth bounding box for learning. Previous works \cite{famnet,bmnet} show that adapting the few-shot object detection methods to the CAC task obtains much poorer performance than the density map-based methods. Previous work \cite{p2pnet} adopts a localization-based scheme for crowd counting and achieves good performance. However, it cannot be directly applied to the CAC task. While \cite{p2pnet} takes image features to predict point locations, we design the HECE and EUQC modules to obtain the correlated query tensor for localization in the CAC task. Moreover, we fully exploit the bounding boxes of exemplars for size supervision and propose scale-aware localization, which achieves improved performance and enables the model to predict the approximate size of an object.  
Compared with state-of-the-art CAC methods, our method can not only achieve superior counting performance (extensively verified in Section \ref{sec:experiment}) but also provide the locations and sizes of objects that are useful for downstream tasks.

\section{Experiments}
\label{sec:experiment}

In this section, we first describe the experimental settings and then verify the effectiveness of the proposed SQLNet by extensive evaluation and comparison with state-of-the-art methods on the CAC benchmarks, along with ablation studies of each module to provide a more comprehensive understanding of the proposed method.

\subsection{Experimental Settings}
\noindent\textbf{Implementation Details.}
Following previous works, we use ResNet50 \cite{resnet,chen2024heterogeneous} (the first 4 blocks) as the feature extractor of our method. 
The output features of the four blocks are used for hierarchical exemplar feature extraction, i.e., $L=4$, and the image features output by the last block is used as the query feature. 
The Exemplars Collaborative Enhancement adopts $L_{c}=1$ layers with a token dimension of 1280 and the feature dimension $d_e=1280$, while the Spatial Correlation has $L_{q}=2$ layers with a token dimension of 1024. For both modules, the hidden dimension is 1024, the number of heads in the MHSA layer is 8 and the dropout is 0.1. the interval number $T$ in for Equifrequent Size Prompt is 20. The Channel Correlation utilizes a Linear-ReLU-Linear-Sigmoid network structure.
The balancing weight $\eta$ for point matching is set as $5e-2$. 
We expect each position on the feature map $\mathcal{F}_r$ to predict 4 nearby points, that is, $N_a=4$, which is sufficient to generate point proposals many more than the ground truth points. In addition, the weights in the loss function are $\gamma=0.5$, $\lambda_{1}=2e-4$, $\lambda_{2}=5e-5$.

Following the learning setup of FamNet \cite{famnet}, we fix the feature extractor and utilize the Adam optimizer with a learning rate of $1e{-5}$ and a batch size of 1. Images are resized to a height of 384, with the width adjusted correspondingly to maintain the original aspect ratio.

\noindent\textbf{Metrics.}
Following previous works \cite{famnet,bmnet}, we use the mean absolute error (MAE) and the root mean square error (RMSE) to measure the performance of the model:

\vspace{-2ex}\begin{align}
MAE&=\frac{1}{M_I}\sum^{M_I}_{i=1}\left|N_i^{P}-N_i^G\right| \\
RMSE&=\sqrt{\frac{1}{M_I}\sum^{M_I}_{i=1}\left|N_i^{P}-N_i^G\right|^2}
\end{align}
where $M_I$ denotes the total number of testing images. $N_i^{P}$ is the number of predicted points with a confidence score larger than 0.5 and $N_i^G$ is the ground truth number of objects for the $i$-th image.

\begin{table}[!t]
\centering
\small
\caption{Quantitative results on CARPK.  The best results are highlighted in bold. }
\label{table:carpk}
\begin{tabular}{cccc}
\hline
 & Method & MAE & RMSE \\ \hline
\multirow{7}{*}{\begin{tabular}[c]{@{}c@{}}Trained on\\ CARPK\end{tabular}} & GMN \cite{gmn} & 7.48 & 9.90 \\
 & FamNet \cite{famnet} & 18.19 & 33.66 \\
 & RCAC \cite{rcac_eccv} & 13.62 & 19.08 \\
 & SPDCN \cite{SPDCN_bmvc} & 10.07 & 14.12 \\
 & BMNet+ \cite{bmnet} & 5.76 & 7.83 \\
 & SAFECount \cite{SAFECount} & 5.33 & 7.04 \\ \cline{2-4} 
 & SQLNet(Ours) & \textbf{4.89}& \textbf{6.55} \\ \hline
\multirow{7}{*}{\begin{tabular}[c]{@{}c@{}}Trained on\\ FSC-147\\ (cross-dataset\\evaluation)\end{tabular}} & GMN \cite{gmn} & 32.92 & 39.88 \\
 & FamNet \cite{famnet} & 28.84 & 44.47 \\
 & RCAC \cite{rcac_eccv} & 17.98 & 24.21 \\
 & SPDCN \cite{SPDCN_bmvc} & 18.15 & 21.61 \\
 & BMNet+ \cite{bmnet} & 10.44 & 13.77 \\
 & SAFECount \cite{SAFECount} & 16.66 & 24.08 \\ \cline{2-4} 
 & SQLNet(Ours) & \textbf{7.66} & \textbf{9.66} \\ \hline
\end{tabular}
\end{table}

\subsection{Comparison with State of the Arts}

\noindent\textbf{FSC-147.}~
FSC-147 \cite{famnet} is a benchmark dataset for the CAC task, which contains 6135 images and a diverse set of 147 object classes. The object count in each image varies widely, ranging from 7 to 3731 objects, with an average count of 56 objects per image. Each object instance is annotated with a dot at its approximate center. In addition, about three object instances are randomly selected as the exemplars of the object class to be counted in each image. The exemplars are also annotated with bounding boxes. The dataset is split into training, validation, and test sets, comprising 3659, 1286 and 1190 images, respectively. The training set has 89 object classes, and both the validation and test sets have 29 disjoint classes, which means FSC-147 is an open-set object counting dataset that the test classes are previously unseen by the model.

\begin{table}[!t]
\centering
\small
\caption{Evaluation results of the average inference efficiency per image on FSC-147. FLOPs is the number of FLoating point OPerations. The units are giga (G) for FLOPs and millisecond (ms) for the processing time.}
\label{table:complexity}
\begin{tabular}{cccc}
\hline
Methods & Paradigm & FLOPs (G)  & Time (ms) \\ 
\hline
Faster-RCNN \cite{fastrcnn} & Detection  & 14.51 & 37.98  \\ 
Mask-RCNN \cite{maskrcnn} & Detection  & 15.54 & 42.97  \\ 
BMNet+ \cite{famnet} & Regression  & 51.6 & 29.97  \\ 
SAFECount \cite{SAFECount} & Regression  & 458.3 & 188.68  \\ 
\hline
SQLNet(Ours) & Localization  & 31.3 & 26.38  \\ 
\hline
\end{tabular}
\end{table}

\begin{table}[!t]
\centering
\small
\caption{Ablation study on the HECE module. \textbf{HEE}, \textbf{ECP} and \textbf{ESP} stand for Hierarchical Exemplar Extraction, Exemplars Collaborative Enhancement, and Equifrequent Size Prompt, respectively.}
\label{table:ablation}
\begin{tabular}{ccccccc}
\hline
\multicolumn{3}{c}{Component} & \multicolumn{2}{c}{Val} & \multicolumn{2}{c}{Test} \\ \hline
 HEE & ECE & ESP  & MAE & RMSE & MAE & RMSE \\ \hline 
 \Checkmark & \XSolidBrush & \XSolidBrush & 15.56 & 58.56 & 16.16 & 114.27 \\ 
 \XSolidBrush & \Checkmark & \XSolidBrush & 15.07 & 55.26 & 15.62 & 96.16 \\ 
 \XSolidBrush & \Checkmark & \Checkmark & 13.61 & 46.59 & 13.04 & 83.23 \\ 
 \Checkmark & \Checkmark & \Checkmark & {\textbf{12.40}} & {\textbf{42.30}} & {\textbf{12.49}} & {\textbf{80.85}} \\ \hline
\end{tabular}
\end{table}

We compare our SQLNet method with the baselines employed in \cite{famnet}, including MAML \cite{maml}, GMN \cite{gmn}, FR \cite{fr} and FSOD \cite{fsod}, and also the state-of-the-art methods FamNet \cite{famnet}, RCAC \cite{rcac_eccv}, BMNet+ \cite{bmnet}, SPDCN \cite{SPDCN_bmvc}, SAFECount \cite{SAFECount} and PseCo \cite{HuangD0ZS24cvpr}.  
As exhibited in Table \ref{table:result}, SQLNet outperforms the state-of-the-art methods in both the MAE and RMSE metrics. For example, compared to the second-best method SPDCN in the MAE metric, our method achieves a drop of 2.19 (15.0\%) and 1.19 (8.8\%) on Val and Test datasets, respectively. 
While compared to the second best method SAFECount in the RMSE metric, our method achieves a drop of 4.90 (10.4\%) and 4.69 (5.5\%)  on Val and Test datasets, respectively.
Furthermore, in contrast to the density regression-based state-of-the-art methods, our localization-based SQLNet is considered more akin to a detection-based approach, which meets the practical demands of a wider range of downstream tasks beyond merely counting but also encounters more difficulties. Nevertheless, our SQLNet still delivers remarkable results. 
The more recent method, PseCo \cite{HuangD0ZS24cvpr}, is a counting approach based on the powerful Segment Anything Model (SAM) \cite{KirillovMRMRGXW23iccv}. It leverages SAM to segment all objects in the image and then employs the CLIP model \cite{CLIP} to match the segmented objects with given examples, thereby completing the classification and counting process. Our method also outperforms PseCo in all metrics.

Qualitative results in multiple scenarios are shown in Figure \ref{fig:compare} for further analysis of our method. The results of the state-of-the-art method SAFECount are also visualized for comparison. 
As can be observed, our SQLNet shows more accurate counting performance than SAEFCount across various scenarios and object classes. 
Although the hotspots in the density map predicted by SAEFCount may provide rough locations of some objects, they are not accurate and mistaken objects are easily introduced with the actual objects missed, partly due to the intrinsic nature of density estimation. 
In contrast, our SQLNet can locate each object well, as verified by its superior counting performance.
Notably, SQLNet surpasses expectations in generating precise bounding boxes, even when exemplars boxes in FSC-147 are less accurately labeled.

\begin{table}[!t]
\centering
\small
\caption{Ablation study on the EUQC module. \textbf{SC} and \textbf{CC} stand for Spatial Correlation and Channel-wise Correlation, respectively.}
\label{table:ablationEUQC}
\centering
\begin{tabular}{cccccc}
\hline
\multicolumn{2}{c}{Component} & \multicolumn{2}{c}{Val} & \multicolumn{2}{c}{Test} \\ \hline
SC & CC   & MAE & RMSE & MAE & RMSE \\ \hline 
\Checkmark & \XSolidBrush & {13.90} & {53.45} & {14.91} & {98.04} \\ 
\Checkmark & \Checkmark & {\textbf{12.40}} & {\textbf{42.30}} & {\textbf{12.49}} & {\textbf{80.85}} \\ \hline
\end{tabular}
\end{table}

\begin{table}[!t]
\centering
\small
\caption{Ablation study on the SAML module. \textbf{SS} stands for Size Supervision.}
\label{table:ablationSAML}
\centering
\begin{tabular}{ccccccccc}
\hline
Component & \multicolumn{2}{c}{Val} & \multicolumn{2}{c}{Test} \\ \hline
 SS   & MAE & RMSE & MAE & RMSE \\ \hline 
 \XSolidBrush & {13.45} & {50.5} & {12.55} & {86.22} \\ 
 \Checkmark & {\textbf{12.40}} & {\textbf{42.30}} & {\textbf{12.49}} & {\textbf{80.85}} \\ \hline
\end{tabular}
\end{table}

\noindent\textbf{Val-COCO and Test-COCO.}~
Following previous works, we also evaluate our model on Val-COCO and Test-COCO datasets. The images in these two datasets are collected from the COCO dataset \cite{Lin2014eccvCOCO}. Val-COCO and Test-COCO contain 277 and 282 images, respectively, and are also subsets of the validation and test sets of FSC-147. These two subnets are commonly used as a separate evaluation benchmark, especially for the comparison with detection-based methods, since COCO is a widely used object detection benchmark. 
As exhibited in Table \ref{table:coco}, our SQLNet approach surpasses all the other compared methods, including the object detectors Faster-RCNN \cite{fastrcnn}, RetinaNet \cite{retinanet} and Mask-RCNN \cite{maskrcnn} that are pre-trained on the COCO benchmark and the state-of-the-art regression-based CAC methods, except for having a slightly higher RMSE value than SAFECount.

\begin{table*}[!ht]
\centering
\small
\caption{Evaluation results of different methods using localization and regression paradigms. $\dag$ denotes the model is modified based on official codes.}\label{tab:LocVsReg}
\begin{tabular}{cccccc}
\hline
\multirow{2}{*}{Paradigm}     & \multirow{2}{*}{Method} & \multicolumn{2}{c}{Val} & \multicolumn{2}{c}{Test} \\ \cline{3-6}
                              &                          & MAE        & RMSE        & MAE        & RMSE         \\ \hline
\multirow{3}{*}{Regression}   & FamNet \cite{famnet}                   & 23.75      & 69.07      & 22.08      & 99.54       \\
                              & BMNet+ \cite{bmnet}                   & 15.74      & 58.53      & 14.62      & \textbf{91.83}       \\
                              & SQLNet with Reg           & \textbf{14.12}      & \textbf{52.15}      & \textbf{13.69}       & 99.20       \\ \hline
\multirow{3}{*}{Localization} & FamNet \cite{famnet} with Loc \cite{p2pnet} $\dag$         & 22.17      & 76.92      & 19.42      & 107.78      \\
                              & BMNet+ \cite{bmnet} with Loc \cite{p2pnet} $\dag$          & 15.28      & 51.72      & 14.63      & 89.36       \\
                              & SQLNet                    & \textbf{12.40}      & \textbf{42.30}      & \textbf{12.49}      & \textbf{80.85}       \\ \hline
\end{tabular}
\end{table*}

\noindent\textbf{CARPK.}~
The CARPK \cite{CARPK} dataset is a car counting benchmark and has been utilized in several CAC works  \cite{famnet,bmnet,SAFECount} to measure model generalization. It contains 1448 images captured from bird's-eye views, with 989 and 459 images as the training and test sets, respectively. The images encompass approximately 90,000 cars and are collected from diverse scenes of four different parking lots. 
Following previous works, we conduct an experiment to evaluate the models trained on the CARPK dataset. We train our model on CARPK in the same way as previous works did (i.e., labeling the centers of bounding boxes as ground truth points and using the same set of 12 exemplars from the training set). The results are reported in Table \ref{table:carpk}. As can be observed, our SQLNet consistently  outperforms all the state-of-the-art methods.

\noindent\textbf{Analysis of Cross-dataset Applicability.}~
We further conduct a cross-dataset evaluation, where the models trained on the FSC-147 dataset are directly evaluated on the CARPK dataset.
As can be observed in the lower part of Table \ref{table:carpk}, our proposed SQLNet outperforms previous state-of-the-art methods by sizable margins in both MAE and RMSE, demonstrating the excellent generalization ability of our method. 
It is worth noting that our method, even without being trained on the CARPK dataset, can achieve better performance than the previous leading methods FamNet, RCAC, and SPDCN that are fine-tuned on CARPK.
This further highlights the superiority of our framework design and its strong generalization ability across different datasets and scenarios.

\begin{figure}[!t]
    \centering
    \includegraphics[width=1.0\linewidth]{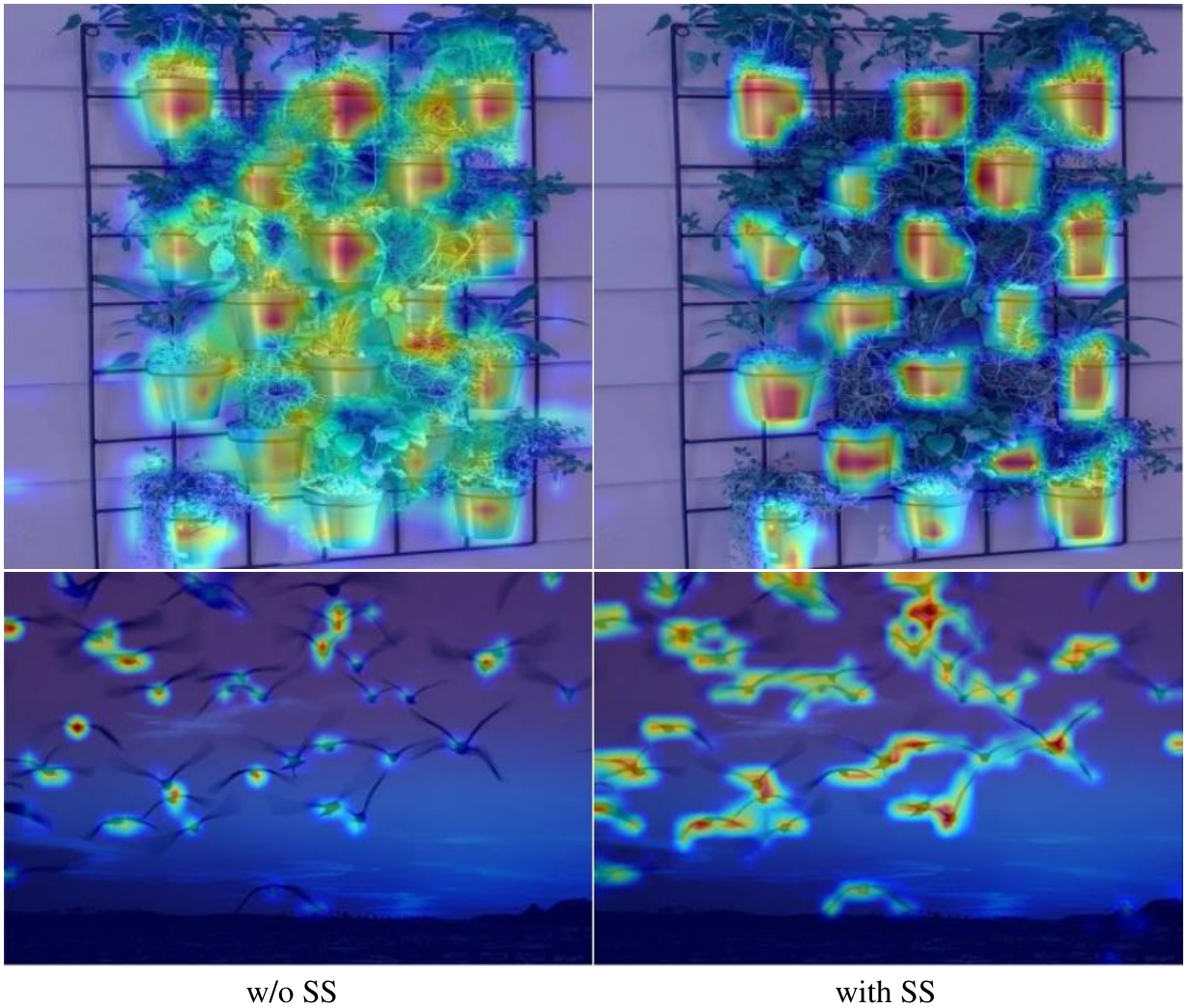}
    \caption{Visualized attention comparison of our model with or without using Size Supervision (SS). The attention maps, visualized using Grad-CAM \cite{gcam}, depict the important regions of the image that the model emphasizes to make predictions.}
    \label{fig:gcam}
\end{figure}

\noindent\textbf{Evaluation on Localization.}~
We adopt the normalized Average Precision (nAP) metric \cite{p2pnet} to evaluate the localization performance, with the parameter $\delta_{nAP}=0.5$. We compare our method with several state-of-the-art object detectors (i.e., Faster-RCNN \cite{fastrcnn}, RetinaNet \cite{retinanet} and Mask-RCNN \cite{maskrcnn}) on the subsets Val-COCO and Test-COCO of FSC-147 for fair comparison, where the object detectors are pre-trained on the COCO dataset. The results are reported in Table \ref{table:coco}. As can be observed, in the counting scenarios, our method outperforms these object detectors by a large margin in nAP, demonstrating the superior localization capability of our method.

\noindent\textbf{Efficiency Comparison.}~
We conduct an experiment to compare our method with existing models on inference efficiency. The execution time is computed on a computer server with an Nvidia GTX 3090 GPU. The average resolution of images in the FSC-147 dataset is $384\times 530$. Table \ref{table:complexity} presents the average computational cost per image in terms of floating point operations (FLOPs) and the processing time. As can be observed, the average processing time of our method is 26.38 milliseconds, lower than the other compared methods. We note that although Faster-RCNN and Mask-RCNN consume less FLOPs, their average processing time is higher than ours. The reason is that they entail a large number of supplementary operations, such as performing extensive non-maximum suppression to obtain region proposals. We further report the training costs of the compared methods. For the detection-based methods Faster-RCNN and Mask-RCNN, we utilize the pre-trained models directly, because fine-tuning them on the CAC dataset is infeasible due to the lack of bounding box annotations.
As revealed by \cite{Hobbhahn2021backwardFLOP,Casson2023transformerflops}, the backward-forward FLOP ratio is about 2:1 for neural network models. Therefore, for each compared method, the average FLOPs consumed in a training iteration is approximately three times that required for inference (shown in Table \ref{table:complexity}).
In terms of training time, both BMNet+ and our method take approximately 3.2 minutes per epoch, whereas SAFECount requires about 13 minutes per epoch.
These comparison results show that our method delivers commendable efficiency in both training and inference.

\subsection{Ablation Study}
\label{section:ablation_study}

We further conduct ablation study experiments on the FSC-147 benchmark and systematically evaluate different variations of our SQLNet model to provide a comprehensive understanding of our method and the modules.

\begin{table}[!t]
\centering
\small
\caption{Evaluation results of our model using different numbers of exemplars in the testing phase. N denotes the number of exemplars.}
\label{tab:number_exemplars}
\begin{tabular}{ccccc}
\hline
N & Val MAE & Val RMSE & Test MAE & Test RMSE \\ \hline
1 & 13.78 & 45.71 & 12.56 & \textbf{77.59} \\
2 & 12.76 & 43.68 & 12.55 & 79.68 \\
3 & \textbf{12.40} & \textbf{42.30} & \textbf{12.49} & 80.85 \\ \hline
\end{tabular}
\end{table}

\noindent\textbf{Ablation on the HECE module.}
The Hierarchical Exemplars Collaborative Enhancement (HECE) module is designed to obtain rich discriminative representations of the target class from the limited exemplars by multi-scale feature collaboration. As stated in Section \ref{sec:Enhancement}, Hierarchical Exemplar Extraction (HEE), Exemplars Collaborative Enhancement (ECE), and Equifrequent Size Prompt (ESP) are introduced for effective model design. We conduct ablation experiments to evaluate how they affect the performance of our model.
Based on the evaluation results reported in Table \ref{table:ablation}, the following observations can be made. 
(i) When only using HEE, i.e., directly extracting multi-scale exemplar features as the class representations, a significant performance drop is witnessed, e.g., an increase of 3.67 and 33.42 in MAE and RMSE on the Test set, respectively.
(ii) When only using ECE with the exemplar features from the last layer, an obvious performance drop is also observed.
(iii) When using ECE and ESP together, ESP can effectively improve the model performance. This suggests that the incorporation of size prompt enables the model to better capture the scale information of exemplars, leading to better discrimination between objects of different sizes.

\begin{table}[!t]
\centering
\small
\caption{Effects of different size prompt settings. }
\label{table:ESE_compare}
\begin{tabular}{ccccc}
\hline
\multirow{2}{*}{\begin{tabular}[c]{@{}c@{}}Size\\ Prompt\end{tabular}} & \multicolumn{2}{c}{Val} & \multicolumn{2}{c}{Test} \\ \cline{2-5} 
 & MAE & RMSE & MAE & RMSE \\ \hline
Uniform & \multicolumn{1}{l}{13.37} & \multicolumn{1}{l}{44.08} & 12.81 & 86.97 \\
ESP & \textbf{12.40} & \textbf{42.30} & \textbf{12.49} & \textbf{80.85} \\ \hline
\end{tabular}
\end{table}

\begin{table}[!t]
\setlength{\tabcolsep}{1.5ex}
\centering
\small
\caption{Evaluation results of our model using different backbones and different pre-trained methods.}
\label{table:compare_arch}
\begin{tabular}{lccccc}
\hline
\multicolumn{1}{c}{\multirow{2}{*}{Backbone}} & \multirow{2}{*}{\begin{tabular}[c]{@{}c@{}}Pre-training\\ Method\end{tabular}} & \multicolumn{2}{c}{Val} & \multicolumn{2}{c}{Test} \\ \cline{3-6} 
\multicolumn{1}{c}{} &  & \multicolumn{1}{l}{MAE} & RMSE & MAE & RMSE \\ \hline
\multirow{2}{*}{ResNet50} & Supervised \cite{resnet} & 14.22 & 50.20 & 13.94 & 93.72 \\
 & Unsupervised \cite{mocov2} & \textbf{12.40} & \textbf{42.30} & \textbf{12.49} & \textbf{80.85} \\ \hline
\multirow{2}{*}{ViT-base} & Supervised \cite{vit} & 21.02 & 77.68 & 19.25 & 122.54 \\
 & Unsupervised \cite{mae} & \textbf{13.39} & \textbf{54.09} & \textbf{14.53} & \textbf{108.09} \\ \hline
\multirow{2}{*}{ViT-large} & Supervised \cite{vit} & 19.89 & 72.88 & 17.46 & 103.40 \\
 & Unsupervised \cite{mae} & \textbf{12.32} &\textbf{51.66} & \textbf{11.37} & \textbf{78.04} \\ \hline
\end{tabular}
\end{table}

\noindent\textbf{Ablation on the EUQC module.}
The Exemplars-Unified Query Correlation (EUQC) module comprises two components, i.e., Spatial Correlation (SC) and Channel-wise Correlation (CC). We conduct an ablation experiment to analyze their impact on the model performance. As shown in Table \ref{table:ablationEUQC}, when the channel-wise correlation is not utilized, there is a significant decrease in model performance on both the Validation and Test sets. This indicates that the inclusion of channel-wise correlation allows the model to better mine the correlation information between image features and class representations.

\noindent\textbf{Ablation on the SAML module.}
The Scale-Aware Multi-head Localization (SAML) module utilizes three heads to predict the confidence, location, and size of each object, and our method exploits a scale-aware localization scheme for learning. Here we conduct an ablation experiment on the size head, i.e., how the size supervision affects the performance of our model.
As shown in Table \ref{table:ablationSAML}, when the size head and size supervision is not utilized, there is a moderate drop in the model performance, which verifies the effectiveness of our scale-aware scheme.
It can also be observed that the incorporation of size supervision yields a more significant performance improvement in RMSE compared to MAE.
The reason may be that, by allowing the model to predict the object size, size supervision guides the model to focus more on the complete object within the specified bounding box. This leads to increased robustness to background changes and more stable prediction results.
To verify this, we employ Grad-CAM \cite{gcam} to visualize the attention maps of our model with and without size supervision. As depicted in Figure \ref{fig:gcam}, the attention map generated by our model with size supervision focuses on the target objects more accurately, while the one without size supervision appears more scattered and influenced by the background. These experiments provide evidence that size supervision in our scheme can enhance model interpretability and prediction stability by guiding the model to prioritize the complete objects.

\begin{table}[!t]
\centering
\small
\caption{Evaluation results of different training data proportions. }
\label{table:data_proportion}
\begin{tabular}{ccccc}
\hline
\multirow{2}{*}{\begin{tabular}[c]{@{}c@{}}Data\\ Proportion\end{tabular}} & \multicolumn{2}{c}{Val} & \multicolumn{2}{c}{Test} \\ \cline{2-5} 
 & MAE & RMSE & MAE & RMSE \\ \hline
10\%                        & 20.44 & 62.70 & 19.61 & 96.04 \\ 
20\%                        & 17.43 & 59.48 & 16.52 & 97.72 \\ 
50\%                        & 15.19 & 52.39 & 14.81 & 95.85 \\ 
100\%                       & \textbf{12.40} & \textbf{42.30} & \textbf{12.49} & \textbf{80.85}    \\ 
\hline
\end{tabular}
\end{table}

\noindent\textbf{Regression VS Localization.}
The front end of our model architecture is designed to fully capture the correlation information between the image and exemplars. To further verify the effectiveness of our novel design, we combine it with a regression head to predict the density map for CAC counting.
As shown in Table \ref{tab:LocVsReg}, by employing a density regression scheme, our ``SQLNet with Reg'' model can outperform the previous best regression-based approaches except for a higher RMSE value on the Test set.
By comparing the ``SQLNet with Reg'' model with the full version of SQLNet, we can observe the latter achieves a notable performance gain, which evidently verifies the effectiveness of our scale-aware localization scheme. 
Moreover, we adapt the regression-based state-of-the-art methods FamNet and BMNet+ by replacing the regression head with the localization scheme from \cite{p2pnet}, obtaining their localization versions. 
As exhibited in Table \ref{tab:LocVsReg}, the localization versions of FamNet and BMNet+ achieve slightly better results than the original ones, but clear advantages are not observed. In contrast, our localization-based SQLNet shows a significant performance improvement, further verifying the effectiveness of our network design. Two reasonable conclusions could be drawn from the results: (i) The localization mechanism can play a critical role in elevating the performance of existing methods for the CAC task. (ii) A good integration of such mechanism is not straightforward and requires some  exquisite designs.

\noindent\textbf{Number of Exemplars.}
We investigate the influence of using different numbers of exemplars on the performance of our SQLNet model during the testing phase, and the results are presented in Table \ref{tab:number_exemplars}. 
By referring to Table \ref{table:result}, we can observe that SQLNet achieves superior results over all the compared methods even when using only one exemplar during testing. 
This is attributed to our comprehensive utilization of exemplar information.
Even with a single exemplar, the model leverages the exemplar features  across different layers to construct a comprehensive class-specific feature bank for discriminative representation enhancement of the current object class, which helps to ensure high performance.

\noindent\textbf{Choice of Size Prompt.}
\label{exp:size_embedding}
Our model exploits Equifrequent Size Prompt (ESP) for the collaborative enhancement of exemplars. Here we compare it with a uniform one, i.e., the range of width and height is divided into uniformly-spaced intervals.
The results presented in Table \ref{table:ESE_compare} demonstrate ESP can effectively improve the performance of our model.

\begin{figure}[!t]
    \centering
    \includegraphics[width=0.98\linewidth]{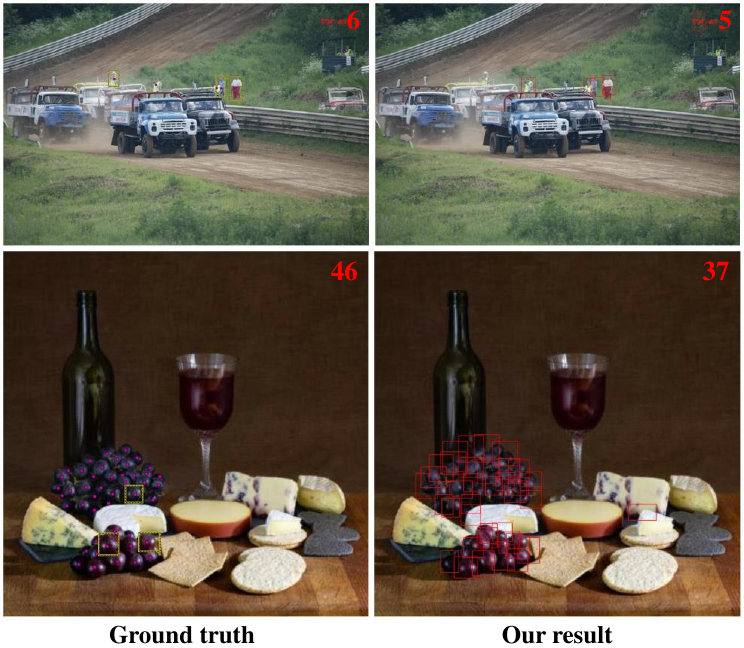}
    \caption{Visualized examples of our results on multi-class images. The top-right corner of an image exhibits the number. }
    \label{fig:multiclass}
\end{figure}

\noindent\textbf{Choice of Feature Extractor.}
\label{sec:feature_extractor}
For implementation, we follow previous works to adopt ResNet50 as our feature extractor. However, it is worth noting that any well-performing backbone network can be utilized as the feature extractor. 
We conduct experiments to evaluate the impact of using different feature extractors on the performance of our SQLNet. 
Specifically, we compare two kinds of well-studied architectures, i.e., the convolution-based ResNet \cite{resnet} and the Transformer-based ViT \cite{vit}.
For ViT, we extract exemplar features from each Transformer layer.
Motivated by previous work BMNet using a backbone with unsupervised pre-training \cite{swav}, we also evaluate the backbones pre-trained with supervised and unsupervised learning methods.  
The evaluation results in Table \ref{table:compare_arch} indicate that the backbone models pre-trained with unsupervised learning outperform the ones with supervised learning. The reason may be that unsupervised pre-training can exploit the intrinsic and more comprehensive feature characteristics from unlabeled data, which is more suitable for the CAC task. 
It can also be observed that using the ResNet50 backbone, our SQLNet performs consistently well in most settings, except slightly worse than that with the unsupervised pre-trained ViT-large backbone in some scenarios. However, ViT-large has an order of magnitude more parameters than ResNet50, resulting in significantly higher computational costs.
Therefore, ResNet50 is considered to be a better choice in practical applications.

\noindent\textbf{Analysis of Limited Training Data.}
The training set of the FSC-147 benchmark contains 3659 images. We conduct an experiment to analyze how our model performs under limited training data. For comparison, we use 10\%, 20\%, 50\% and 100\% of the training data to train our model, respectively. The evaluation results are reported in Table \ref{table:data_proportion}. 
When the data proportion is 10\%, our model exhibits a sharp MAE increase of 8.04 and and 7.12 on Val and Test sets, respectively.
As the data proportion increases, our model exhibits a sustained improvement in performance. It implies that the quantity of training data has a significant impact on the performance of our model.
In addition, a comparison with the results presented in Table \ref{table:result} reveals that our model trained with 10\% of the training data can outperform the leading methods FamNet \cite{famnet} and RCAC \cite{rcac_eccv}.
It further demonstrates the superiority of our framework design. Our framework better exploits the training data in terms of feature extraction, feature interaction and supervision through the three key modules. We plan to explore more theoretical analyses on it in the future.

\noindent\textbf{Analysis on Multi-class Images.}
Images in the FSC-147 dataset mainly contain objects from a single class for counting. To show our method's performance on images containing objects from multiple classes, we have selected some appropriate images from existing datasets (e.g., COCO). Several example results are visualized in Figure \ref{fig:multiclass}. It should be noted that the results are acceptable, but not good enough. We believe that an important reason is the insufficiency in both the quantity and diversity of images containing objects of multiple classes in the training set. Consequently, in the future, we plan to collect and annotate richer and more diverse data of this kind for model training and more comprehensive evaluation.

\begin{figure}[!t]
    \centering
    \includegraphics[width=0.98\linewidth]{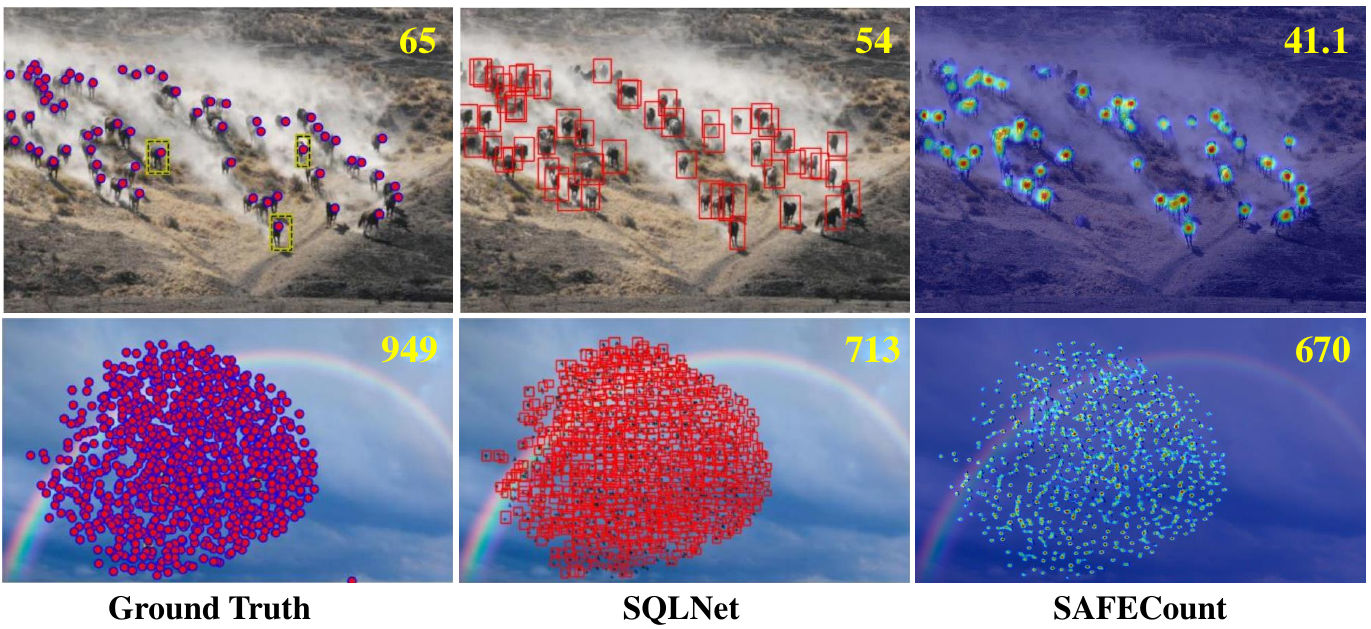}
    \caption{Visualized examples of some bad cases. The top-right corner of an image exhibits the number. }
    \label{fig:badcases}
\end{figure}

\noindent\textbf{Analysis of Bad Cases.} We further analyze some bad cases where  our method shows a higher rate of missed predictions of target objects. As shown in the first row of Figure \ref{fig:badcases}, the dust raised by the running horses in the image causes substantial occlusion, makes it more difficult for our method to discover the horses. In the second row, where the birds are extremely small and densely crowded, our method encounters significant challenges and misses over 200 target objects. However, we also observe that our method still outperforms the leading method SAFECount in these scenarios.

\section{Conclusion}
\label{sec:conclusion}

In this work, we present a novel approach termed SQLNet to address the class-agnostic counting (CAC) task by fully exploring the scales of exemplars in both the query and localization stages of our framework.
In the query stage, to obtain sufficient correlation information between the query image and the exemplars, our SQLNet introduces novel architectures to exploit collaborative enhancement of multi-scale exemplars and perform their interactions with the query features in an exemplars-unified manner. 
Further, a scale-aware localization paradigm is introduced, enabling our SQLNet to achieve excellent counting performance by accurately locating each object and predicting its approximate size in the localization stage. 
Extensive experiments on multiple benchmarks demonstrate that SQLNet achieves state-of-the-art counting performance by a considerable margin over previous leading methods. Meanwhile, it also shows excellent performance in localization and bounding box generation, offering a practical solution for downstream tasks beyond counting.


\bibliographystyle{IEEEtran}
\bibliography{reference}

\end{document}